\documentclass[letterpaper, 10 pt, conference]{ieeeconf}  

\IEEEoverridecommandlockouts                              

\usepackage{cite}
\usepackage{amsmath,amssymb,amsfonts}
\usepackage{algorithm}
\usepackage{algpseudocode}
\usepackage{graphicx}
\usepackage{hyperref}
\usepackage{textcomp}
\usepackage{xcolor}
\usepackage{subcaption}
\usepackage{caption}
\usepackage{multicol}
\usepackage{array, makecell} %
\usepackage{flushend}
\usepackage{soul}
\let\labelindent\relax
\usepackage{enumitem}
\usepackage{balance}
\usepackage[cal=euler]{mathalfa}

\title{\LARGE \bf
Contact Energy Based Hindsight Experience Prioritization}

\author{Erdi Sayar$^{1}$, Zhenshan Bing$^{1,*}$, Carlo D'Eramo$^{2,3,4}$, Ozgur S. Oguz$^{5}$, Alois Knoll$^{1}$
\thanks{$^{1}$Technical University of Munich, erdi.sayar@tum.de, \{bing, knoll\}@in.tum.de}%
\thanks{$^{2}$University of Würzburg, carlo.deramo@uni-wuerzburg.de}%
\thanks{$^{3}$TU Darmstadt, carlo.deramo@tu-darmstadt.de}%
\thanks{$^{4}$Hessian.AI, The Hessian Center for Artificial Intelligence}%
\thanks{$^{5}$Bilkent University, ozgur@cs.bilkent.edu.tr}
\thanks{Corresponding author: Zhenshan Bing}
}

\begin{document}

\maketitle
\thispagestyle{empty}
\pagestyle{empty}

\begin{abstract}
Multi-goal robot manipulation tasks with sparse rewards are difficult for reinforcement learning (RL) algorithms due to the inefficiency in collecting successful experiences. Recent algorithms such as Hindsight Experience Replay (HER) expedite learning by taking advantage of failed trajectories and replacing the desired goal with one of the achieved states so that any failed trajectory can be utilized as a contribution to learning. However, HER uniformly chooses failed trajectories, without taking into account which ones might be the most valuable for learning.
In this paper, we address this problem and propose a novel approach Contact Energy Based Prioritization~(CEBP) to select the samples from the replay buffer based on rich information due to contact, leveraging the touch sensors in the gripper of the robot and object displacement. Our prioritization scheme favors sampling of contact-rich experiences, which are arguably the ones providing the largest amount of information. We evaluate our proposed approach on various sparse reward robotic tasks and compare it with the state-of-the-art methods. We show that our method surpasses or performs on par with those methods on robot manipulation tasks. 
Finally, we deploy the trained policy from our method to a real Franka robot for a pick-and-place task. We observe that the robot can solve the task successfully.
The videos and code are publicly available at: {\href{https://erdiphd.github.io/HER_force/}{https://erdiphd.github.io/HER\_force/}}.
\end{abstract}

\section{INTRODUCTION}
Reinforcement learning (RL) enables an agent to learn optimal behavior through trial-and-error interactions with an environment \cite{kaelbling1996reinforcement,bing2020energy,bing2022simulation}. Deep RL combined with neural networks has drawn attention to sequential decision-making problems due to its success in exceeding human-level performance in playing Atari games \cite{mnih2013playing}, \cite{mnih2015human}, beating Go master \cite{silver2016mastering} and learning to achieve robotic tasks \cite{ng2006autonomous}, \cite{9772990}, \cite{9466373}.

One of the biggest challenges in robotic tasks is to design an effective reward function \cite{ng1999policy}, which requires domain knowledge and the ability to examine tasks in advance.
However, defining reward functions is not trivial for certain manipulation and control tasks. Therefore, a binary reward, which indicates whether the task was accomplished successfully or not, is often used for developing RL algorithms. 
Unfortunately, binary rewards result in a sparse reward signal due to the insufficient amount of successful experiences.
Hindsight experience replay (HER)~\cite{AndrychowiczWRS17} addresses the problem of sparse reward by substituting desired goals with the achieved goals sampled from failed trajectories (i.e., the sequence of interactions between the agent and the environment).
In essence, failed episodes are transformed into successful ones.
Since HER samples achieved goals uniformly from the visited states, it does not take into account which visited states are most beneficial for learning, resulting in sample inefficiency.
However, prioritization of trajectories may alleviate this problem.
For instance, Prioritized Experience Replay (PER)~\cite{schaul2015prioritized} prioritizes trajectories based on temporal difference error, Maximum Entropy-based Prioritization (MEP)~\cite{zhao2019maximum} leverages entropy, and Energy-Based Prioritization (EBP)~\cite{zhao2018energy} utilizes object energy. However, none of these methods incorporates contact feedback.

Another significant challenge in robotics is the uncertainty during object grasping.
Incomplete or noisy observations of object pose or shape contribute to this uncertainty, making it harder for robots to perform stable grasps.
Contact feedback is essential during manipulation and grip acquisition, whereas visual feedback is crucial for determining the grasp position.
To address this challenge, several studies have explored the use of touch sensors to enhance robotic grasping. Merzic et al.\ \cite{merzic2018leveraging}  demonstrated the importance of contact feedback in making grasps more stable using a multi-fingered hand with contact sensing.
Wang et al.\ \cite{wang2021swingbot} developed a robot system that uses tactile sensing to determine the physical characteristics of grasping objects. In the realm of RL for manipulation tasks, touch information has also been shown to improve the performance of algorithms.
Melnik et al.\ \cite{melnik2019tactile} analyzed the learning performance of deep RL algorithms for dexterous in-hand object manipulation tasks with and without touch sensors and found that touch information can improve sampling efficiency and overall performance.
Vulin et al.\ \cite{Vulin_2021} used a touch sensor in the end effector of a robot and a touch-based intrinsic reward function (i.e., a contact feedback-based reward) to improve learning progress for multi-goal RL manipulation tasks. However, they introduced a contact-based reward function in addition to the binary reward and a threshold value to restrict cumulative contact forces to two empirically determined discrete values while disregarding discrepancies between contact force values.

We observe that for manipulation tasks, the robot needs to exert force and change the object's position in order to complete the tasks successfully.
Therefore, in our approach, we further extended prior research, EBP~\cite{zhao2018energy} and CPER~\cite{Vulin_2021}, by incorporating object displacement with a touch sensor while exclusively relying on the binary reward signal without the need for additional reward functions. Additionally, we have addressed the uniform sampling strategy of HER by introducing a new replay buffer prioritization method based on touch sensors. Finally, we have implemented Sim2Real transfer.

The main contributions of this paper are the following:

\begin{itemize}[leftmargin=*]
\item Incorporating touch sensors in the robot's end effector to explore the effectiveness of contact in robotic tasks with only relying on the binary sparse reward signal.

\item Introducing a novel replay buffer prioritization scheme Contact Energy Based Prioritization (CEBP) by computing \textit{contact energy}. This leverages touch feedback and displacement of the object. Furthermore, a sigmoid function is used for smoothing contact energy within a range, instead of restricting them to discrete values as in the CPER benchmark. We also conducted an ablation study that examines the impact of using sigmoid function with different parameters.

\item We compare our approach with different benchmarks in 7-DOF Fetch Robotic-arm MuJoCo simulation environment \cite{plappert2018multigoal}. Furthermore, we implemented Sim2Real transfer and tested our policy on a Franka robot to demonstrate the applicability of the proposed prioritization scheme.
\end{itemize}



\begin{figure*}[t!]
	\centering
	\begin{subfigure}[t]{.23\textwidth}
		\centering
		\includegraphics[width=1\linewidth]{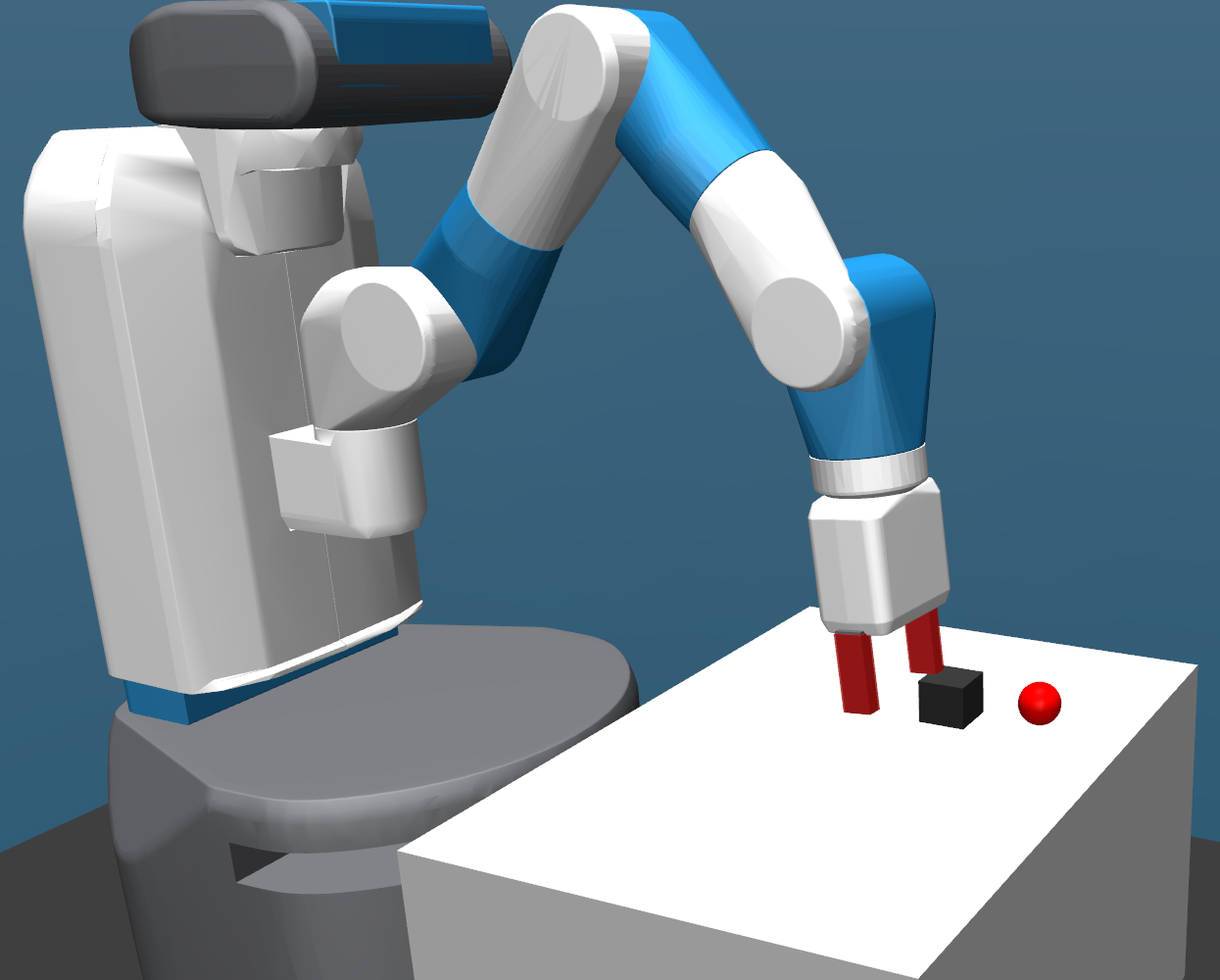}
		\caption{Touch sensors on the left and right side of the gripper.}
		\label{fig:fetchpickandplace_force_sensor}
	\end{subfigure}
	\begin{subfigure}[t]{.23\textwidth}
		\centering
		\includegraphics[width=1\linewidth]{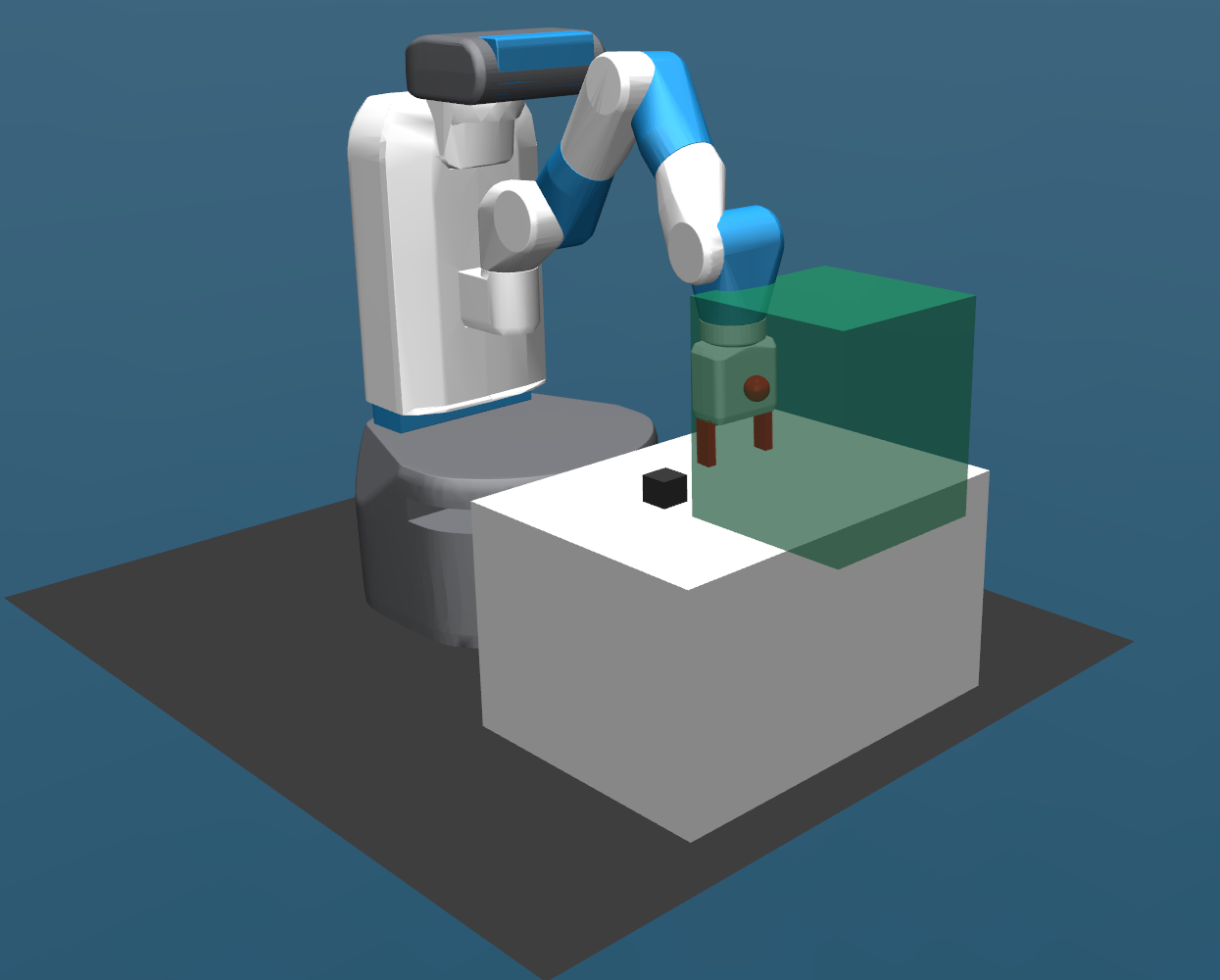}
		\caption{FetchPickAndPlace}
		\label{fig:fetch_pick_and_place}
	\end{subfigure}
	\begin{subfigure}[t]{.23\textwidth}
		\centering
		\includegraphics[width=1\linewidth]{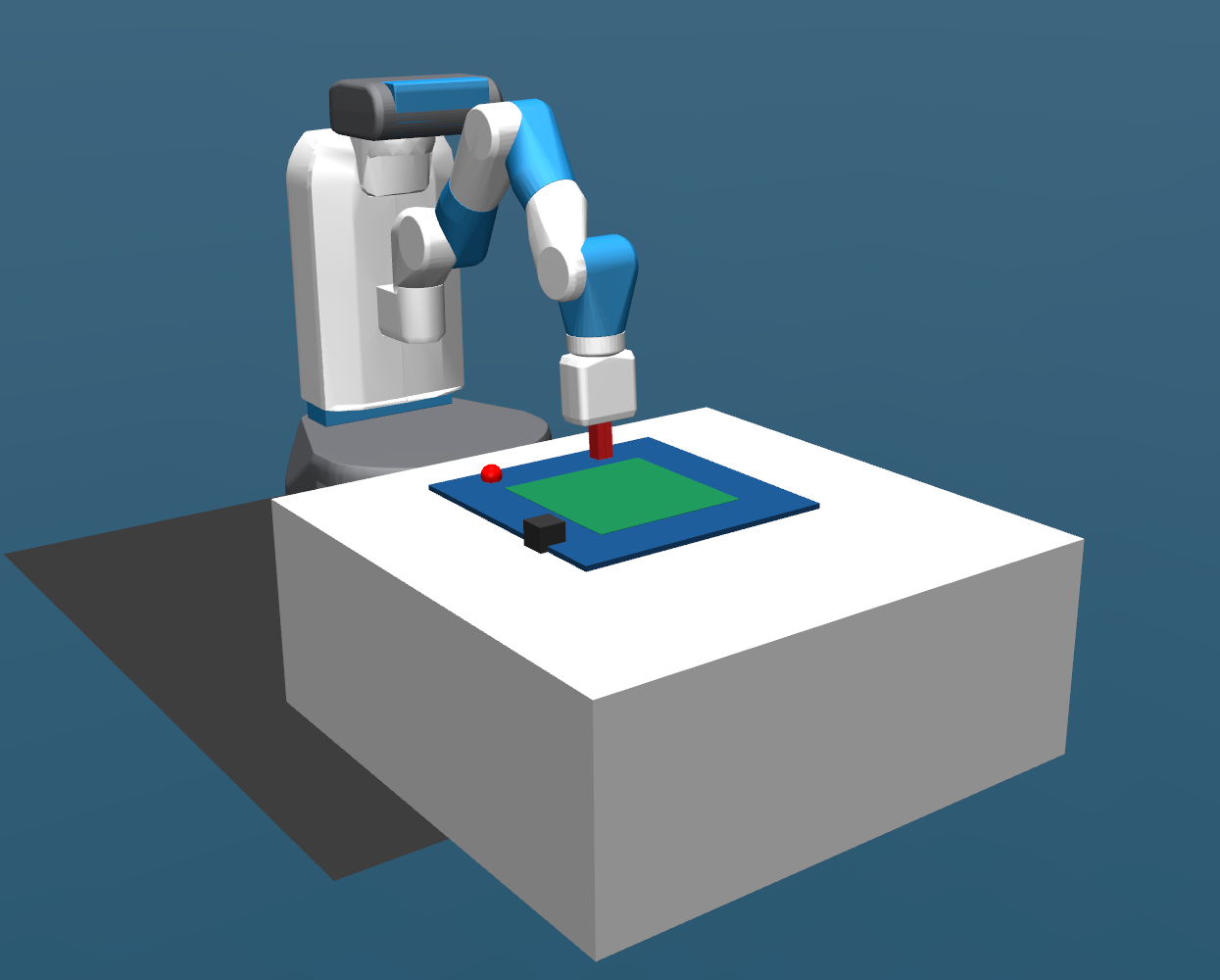}
		\caption{FetchPush}
		\label{fig:fetch_push}
	\end{subfigure}
	\begin{subfigure}[t]{.23\textwidth}
		\centering
		\includegraphics[width=1\linewidth]{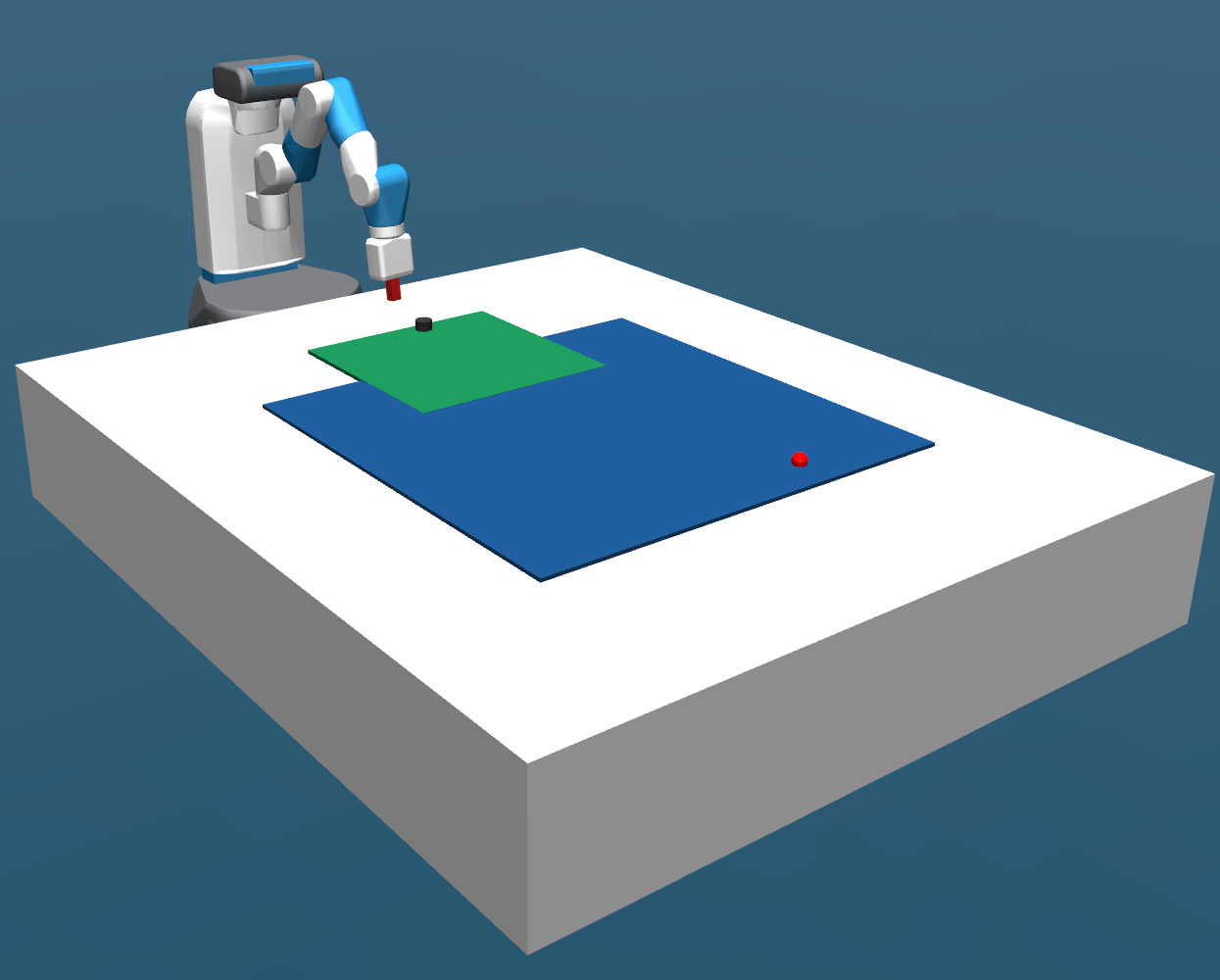}
		\caption{FetchSlide}
		\label{fig:fetch_slide}
	\end{subfigure}
	\caption{Overview of the robotic manipulation benchmark tasks. }
       \label{fig:benchmark_env}
\end{figure*}

\section{RELATED WORK}

\subsection{Tactile Feedback}
Tactile sensor provides local information about contact regions.
Different type of tactile sensors such as capacitive\cite{li2016flexible}, resistive\cite{weiss2005working} and optical-based \cite{donlon2018gelslim} have been made in order to use for various applications such as object recognition\cite{kaboli2019tactile}, slip detection\cite{james2018slip}, pose estimation\cite{bimbo2016hand}.
\color{black}
Tactile sensors have been shown to improve grasping performance and are promising for manipulation tasks \cite{romano2011human}. Manipulation tasks are mostly challenging tasks even with vision-based methods due to occlusion and poor lighting situations. Instead, tactile sensing can provide rich information from multiple contact points \cite{mohsenphdthesis}. These works \cite{van2015learning}, \cite{melnik2019tactile},\cite{koenig2021tactile} have proven that tactile feedback can improve the learning performance.

The works \cite{melnik2019tactile} and \cite{Vulin_2021} inspired us to carry out this research. In the first work \cite{melnik2019tactile} Shadow-Dexterous-Hand are equipped with force sensors in different regions of the hand.
As we explained our difference in the introduction section, in the second work \cite{Vulin_2021}, a touch-dependent intrinsic reward is added to the binary reward as soon as the total amount of touch value is greater than an empirically found threshold.

\subsection{Energy-Based Hindsight Experience Prioritization}
Zhao et al.\ \cite{zhao2018energy} take into account the total trajectory energy to prioritize the experiences. The total trajectory energy consists of the potential energy, the kinetic energy, and the rotational energy of the object, and it is calculated at every timestep. The motivation for this work is that prioritizing experiences that have high trajectory energy is a more efficient approach for reinforcement learning in robotics because the robot must change the energy of an object in order to complete a task. Generally, the difficulty of a task is directly proportional to the amount of work required from the robot. Therefore, the trajectory energy of the object is used to estimate the difficulty of the task. Consequently, prioritizing experiences based on energy levels enhances the efficiency of the learning process.

\subsection{Prioritized Experience Replay}
Schaul et al.\ \cite{schaul2015prioritized} examine how prioritizing certain transitions during experience replay can enhance its efficiency and effectiveness compared to replaying all transitions uniformly. The main concept of this study is that an RL agent can learn better from some transitions than others. Prioritized experience replay (PER) prioritizes experiences according to temporal difference (TD) error which is the discrepancy between the actual value and next-step bootstrap. In other words, TD-error indicates how much the RL agent needs to adjust its predictions to better approximate the true value function. An extension of this approach to multi-task RL is presented in \cite{d2022prioritized}. Prioritizing transitions with higher TD errors means that the agent will focus more on learning from the experiences that it had trouble predicting accurately, which could lead to more effective learning and faster convergence to an optimal policy.

\subsection{Maximum Entropy-based Prioritization}
Maximum Entropy-based Prioritization~\cite{zhao2019maximum} improves the sample efficiency by combining maximum entropy distribution and multi-goal RL. It encourages agents to traverse diverse goal-state trajectories while maximizing expected returns. Sampling the trajectories from a higher entropy distribution helps the agent to assign priority to those under-represented goals. The underlying idea is to prioritize the learning of those goals that are less frequently represented but have a relatively higher value in terms of what can be learned from them.

\subsection{Hindsight Experience Replay}
In environments with sparse rewards, it may take a long time to receive reward signals even after completing several sub-goals. As a result, the agent cannot gather sufficient rewards to reinforce its actions and cannot learn \cite{zai2020deep}. Andrychowicz et al.\ \cite{AndrychowiczWRS17} developed an effective method to address the sparse reward issue by replacing the desired goals with the achieved goals from visited states which can be seen as a form of implicit curriculum learning. Because the hindsight goals in HER are selected uniformly from all visited states in the replay buffer, hindsight goals are mostly distributed around the initial state. If desired goals are far away from the initial states, HER cannot solve these tasks effectively as no reward signal is provided.

\section{BACKGROUND}
Multi-goal RL can be modeled as a goal-oriented Markov Decision Process (MDP) with continuous state and action spaces: $\langle \mathcal{S},\mathcal{A},\mathcal{G},\mathcal{T},\mathcal{R},p,\mathcal{\gamma} \rangle$ where $\mathcal{S}$ is a continuous state space, $\mathcal{A}$  is a continuous action space, $ \mathcal{G}$ is a set of goals, $ \mathcal{T}$ is the unknown transition function, $R(s,g)$ denotes the immediate reward obtained by an agent upon reaching state  $ s \in \mathcal{S}$, $p(s_{0},g)$ is a joint probability distribution over initial states and desired goals and $ \gamma \in [0,1] $ is a discount factor.
The reward function is defined as
\begin{equation}
    \label{eqn:reward_function}
    R(s,g) = \mathbf{1}[ \lVert \phi\left(s\right) - g \rVert_{2} \leq \epsilon] - 1.
\end{equation}
where $\mathbf{1}$ is the indicator function, $\phi$ is a function that maps states to goal representations and $\epsilon$ is a fixed threshold. The agent aims to determine the optimal policy $\pi$ that maximizes the expected cumulative reward which can be estimated by the value function. The idea of the universal value function is to integrate goals into value function and generalize it over the goals. $V^{\pi}(s,g)$ is defined as a goal-based value function under a policy $\mathcal{\pi}$ for any given goal $g$ for all state $s \in S $~\cite{schaul2015universal}.

Within this framework, the learning problem can be represented as an RL problem that seeks a policy $\mathcal{\pi}: \mathcal{S} \times \mathcal{G} \rightarrow \mathcal{A} $ such that the expected discounted sum of rewards is maximized for any given goal.
\setcounter{footnote}{1}
\footnotetext{Benchmark tasks are used exactly as in 
\url{https://github.com/nikolavulin/tactile_intrinsic_motivation/tree/master/Gym}~\cite{Vulin_2021}.}

\begin{figure*}[t!]
	\centering
	\begin{subfigure}[t]{.32\textwidth}
		\centering
		\includegraphics[width=1\linewidth]{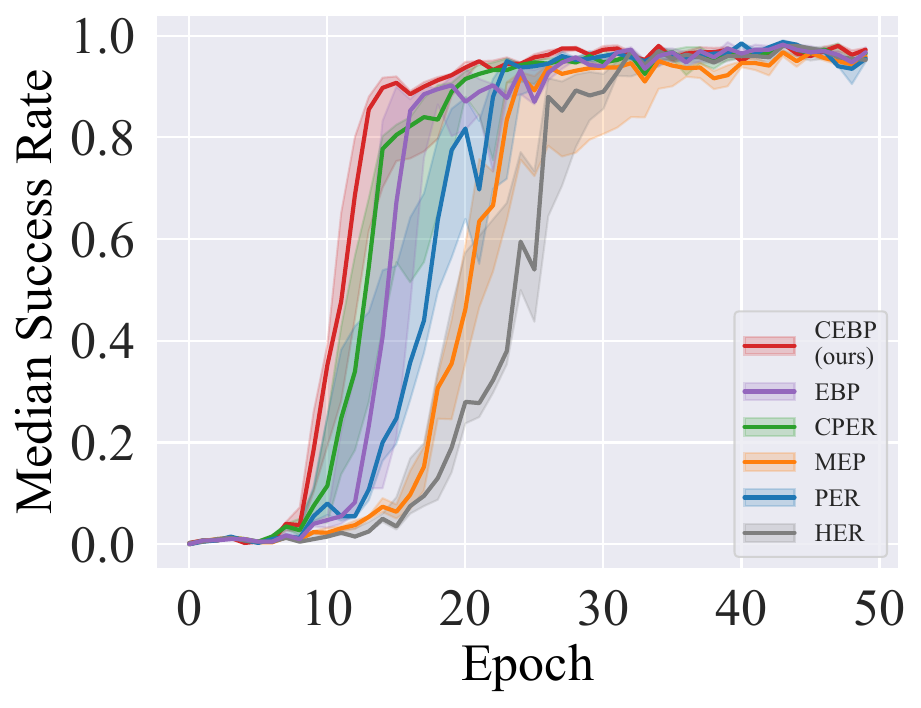}
		\caption{FetchPickAndPlace}
        \label{fig:result_fetch_pick_and_place}
	\end{subfigure}
	\begin{subfigure}[t]{.32\textwidth}
		\centering
		\includegraphics[width=1\linewidth]{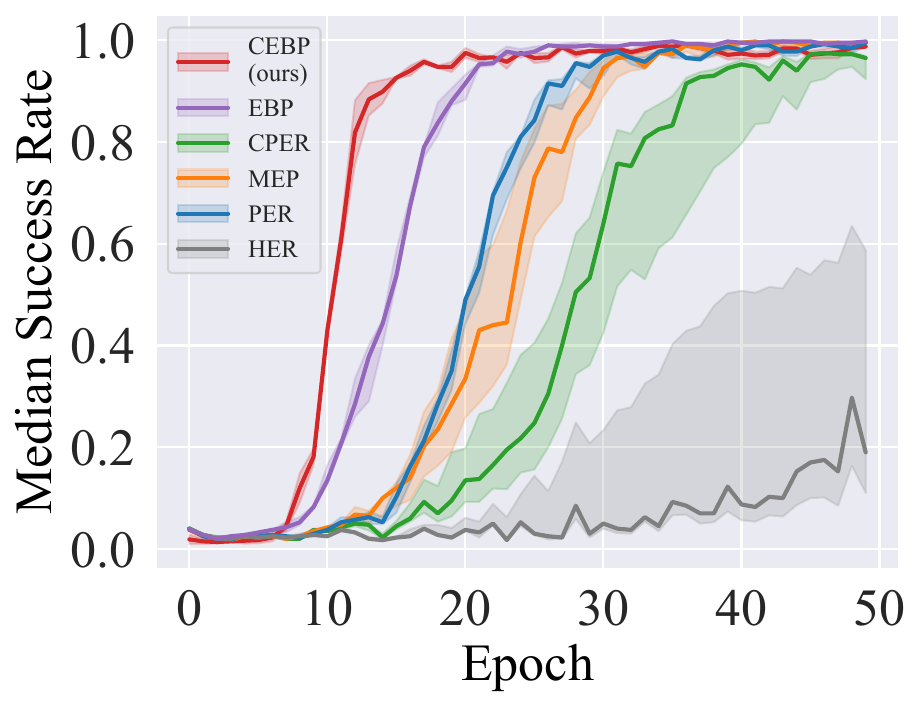}
		 \caption{FetchPush}
            \label{fig:result_fetch_push}
	\end{subfigure}
	\begin{subfigure}[t]{.32\textwidth}
		\centering
		\includegraphics[width=1\linewidth]{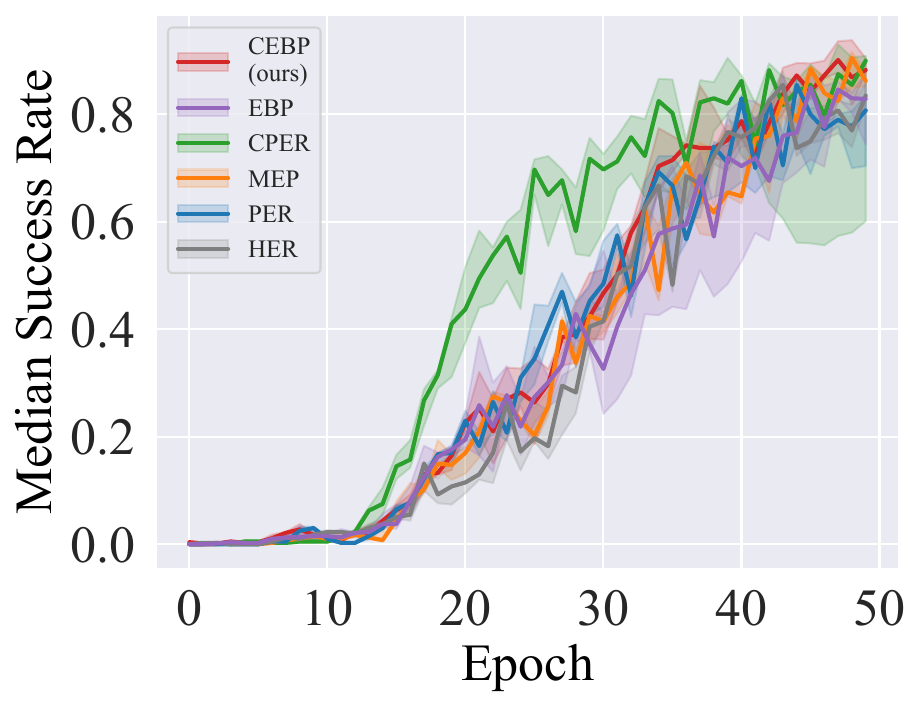}
		 \caption{FetchSlide}
        \label{fig:result_fetch_slide}
	\end{subfigure}
	\caption{Median success rate for all three Fetch tasks. The average success rate (line) and interquartile range (shaded) are shown with training 5 random seeds.  \protect\footnotemark }
       \label{fig:result_benchmark_env}
\end{figure*}
\footnotetext{All methods are trained using the same sparse reward setting.}

\section{METHODOLOGY}
When we train robots to perform manipulation tasks, the sparse reward signal can hinder the agent's learning process by making it difficult to discern which actions lead to positive outcomes, resulting in slow or inefficient policy learning.
HER leverages failed experiences by replacing the desired goal with the achieved goal of the failed experiences. With this modification, any failed experience returns a nonnegative reward.
\color{black}
However, if the initial position of the robot is far from the desired position, hindsight goals generated by HER may not be effective in solving the task because they are sampled randomly from the visited states, which are mostly distributed around the initial state. Consequently, experiences are not regarded based on their importance for learning, which leads to sample inefficiency.

As we are interested in robot manipulation tasks, we leverage the touch sensor and prioritize experiences based on the contact forces occurring between the gripper and the object during the interaction, as well as the object displacement. The contact forces can be used to prioritize experiences that involve more interaction with the gripper, indicating that the robot is indeed manipulating the object. Nevertheless, relying solely on contact force as a criterion for prioritization might be insufficient or even misleading as the gripper may engage with the object in impractical ways, such as exerting downward pressure onto the object's surface. In such cases, the touch sensor detects substantial forces, yet the experiences resulting from the robot's actions fail to yield useful outcomes in manipulating objects. Moreover, these experiences could inadvertently lead such a prioritization algorithm to overemphasize them in the learning process, as larger contact forces would be sampled with a higher probability of learning. 

Additionally, experiences involving collisions between the robot and other objects in its vicinity, like a table, tend to yield unfavorable results. 
Consequently, introducing object displacement as a complementary consideration to contact forces serves to eliminate the prioritization of infeasible scenarios. 
In addition, to diminish the adversarial effect of collisions due to high forces, we employ a sigmoid function to smooth the contact energy, i.e., the extreme forces are saturated.
Furthermore, unlike CPER~\cite{Vulin_2021}, where cumulative, continuous contact forces are converted only to two discrete values \{$1$, $\lambda$\}, thereby dismissing the significance of contact forces except for these predefined values in the set, the sigmoid function allows us to prioritize experiences based on a continuous range.
In the ablation study, we investigated the impact of the sigmoid function with different parameters on the learning process.

\subsection{Contact Energy Prioritization}
\label{sec:contact_energy_replay_buffer}
The touch sensors, as shown in Fig. \ref{fig:fetchpickandplace_force_sensor}, are used on the gripper's left and right sides to measure contact forces. The output of the touch sensor is a non-negative scalar value, which is calculated by summing up the normal forces from all the contact points. Then, contact forces from the left and right sides are summed up into a single value. Next, in each time step, we multiply these contact forces by the Euclidian norm of the object's displacement and calculate the contact energy as 
\begin{equation}
    c(e,t) = \sum_{i=0}^{t}  \left( {f}_{\textrm{left}}\left(e,i \right) +{f}_{ \textrm{right}}\left(e,i\right)  \right) \cdot \lvert \lvert\Delta {d}_{c}(i) \rvert \rvert_{2}.
    \label{eq:contact_energy}
\end{equation}
where $e$ is episode number and $\lvert \lvert\Delta {d}_c(i) \rvert \rvert_{2}$ is the Euclidian distance of the object's displacement between current $i$ and previous timestep $ i -1 $.

The touch sensor relies on contact forces, which can be large during exploration when the gripper collides with objects around it, such as a table. The higher the robot's velocity when it collides with objects, the higher the contact force values will be. However, these experiences with large contact forces are mostly failed trajectories and lack valuable information. As a result, they mislead our replay buffer prioritization algorithm into prioritizing them for learning, as experiences with high contact forces are sampled with higher probability. To address this issue, we are using a sigmoid function to smooth large contact values due to collision.

\begin{equation}
    \sigma(x) = \frac{k}{1 + e^{-xT}}.
\end{equation}
where $k$ is the scale and $T$ is the temperature.
\begin{equation}
    \Tilde{c}(e,t) = \sigma(c(e,t)).
\end{equation}
where $\Tilde{c}$ is the smoothed contact energy.

Then the probability distribution of the episodes in equation~\eqref{eq:p_episode_based_on_contact} is calculated based on their contact-energy rich information, and a mini-batch $\mathcal{B}$ is sampled from the replay buffer $\mathcal{R}$ according to the probability distribution $p_{\textrm{episode}}(e)$
\begin{equation}
    p_{\textrm{episode}}(e) = \frac{\sum_{t=0}^{\mathcal{T}}\Tilde{c}(e,t)}{\sum_{i_{e}=0}^{e} \sum_{t=0}^{\mathcal{T}}\Tilde{c}(i_{e},t)}.
    \label{eq:p_episode_based_on_contact}
\end{equation}

Prioritizing the experiences according to the given methods above introduces bias because it changes the distribution, thereby changing the ultimate solution that the estimates will converge to. This bias can be removed by using the importance-sampling (IS) weights given in equation~\eqref{eq:bias_correction} 
\begin{equation}
\label{eq:bias_correction}
w_{i} = \left( N \cdot p_{\textrm{episode}}(e) \right)^{-\beta}.
\end{equation}
Then, these weights are multiplied with TD error  $\delta$ and the Q-value is updated. 
For stability reasons, weights are normalized by $1/ \max(w)$ \cite{schaul2015prioritized}.

Algorithm \ref{alg:alg} outlines our method. It can be used in combination with any off-policy algorithm, and we choose the Deep Deterministic Policy Gradients (DDPG)~\cite{lillicrap2015continuous} algorithm for comparing the results in section \ref{sec:experiment}. Then, a desired goal is sampled, and trajectories are collected by running the policy. If the end state of the episode is chosen as a hindsight goal, then the desired goal $g$ should be replaced with this achieved goal, and the reward function should be recalculated. The probability distribution $ p_{\textrm{episode}}(e)$ is calculated using equation~\eqref{eq:p_episode_based_on_contact} and a mini-batch is sampled according to $ p_{\textrm{episode}}(e)$ from the replay buffer $\mathcal{R}$. 
The importance sampling weight (eq.~\ref{eq:bias_correction}) and the TD-error are then calculated for each transition sampled from the mini batch. Finally, the weights are updated.


\begin{figure*}[t!]
	\centering
	\begin{subfigure}[t]{.32\textwidth}
		\centering
		\includegraphics[width=1\linewidth]{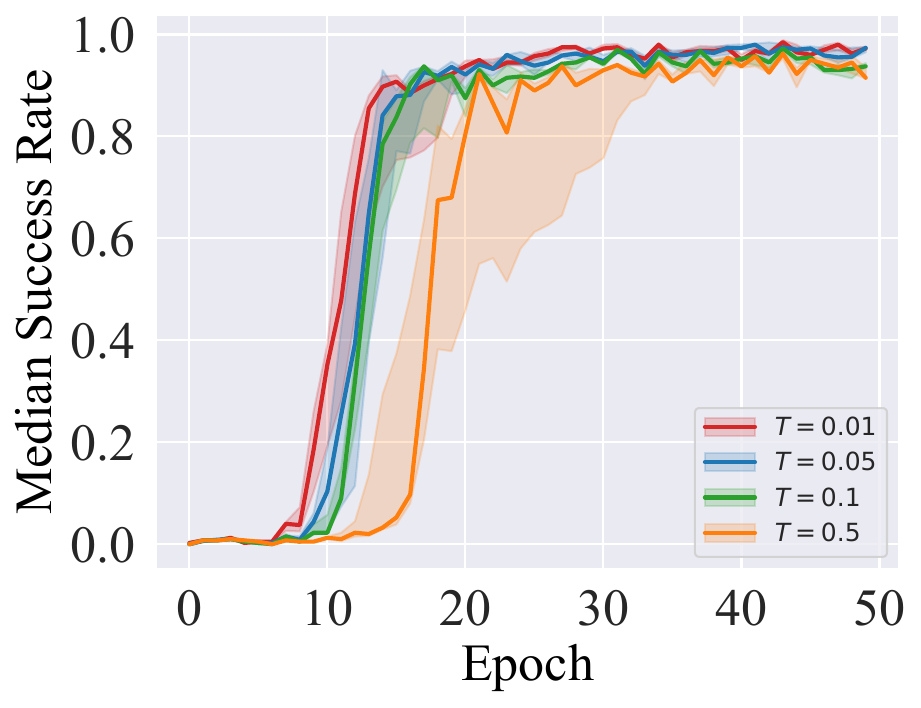}
        \caption{FetchPickAndPlace}
        \label{fig:ablation_fetch_pick_and_place}
	\end{subfigure}
	\begin{subfigure}[t]{.32\textwidth}
		\centering
		\includegraphics[width=1\linewidth]{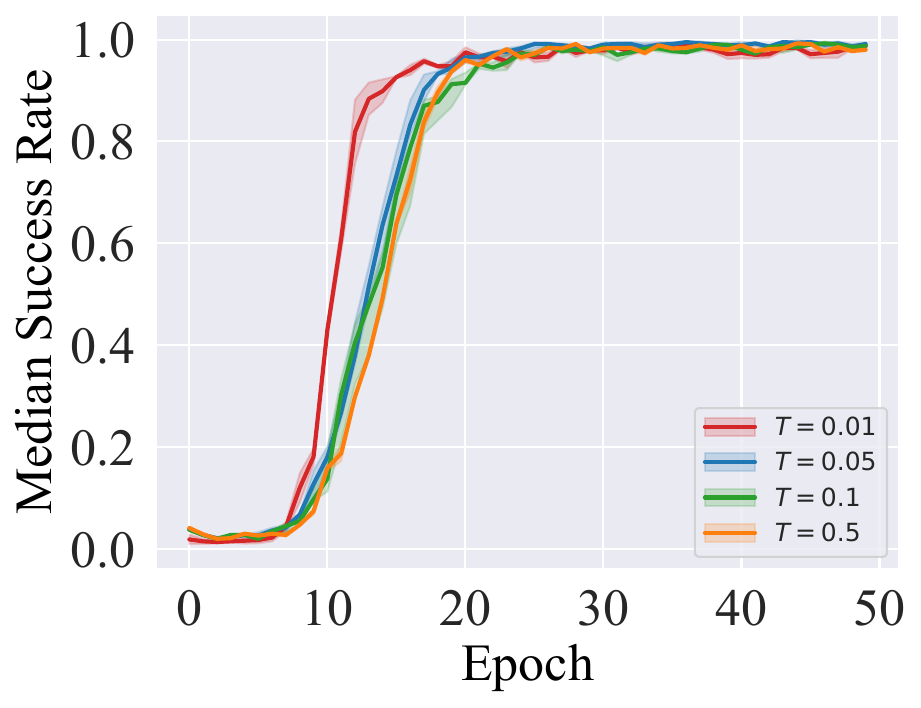}
        \caption{FetchPush}
        \label{fig:ablaion_fetch_push}
	\end{subfigure}
	\begin{subfigure}[t]{.32\textwidth}
		\centering
		\includegraphics[width=1\linewidth]{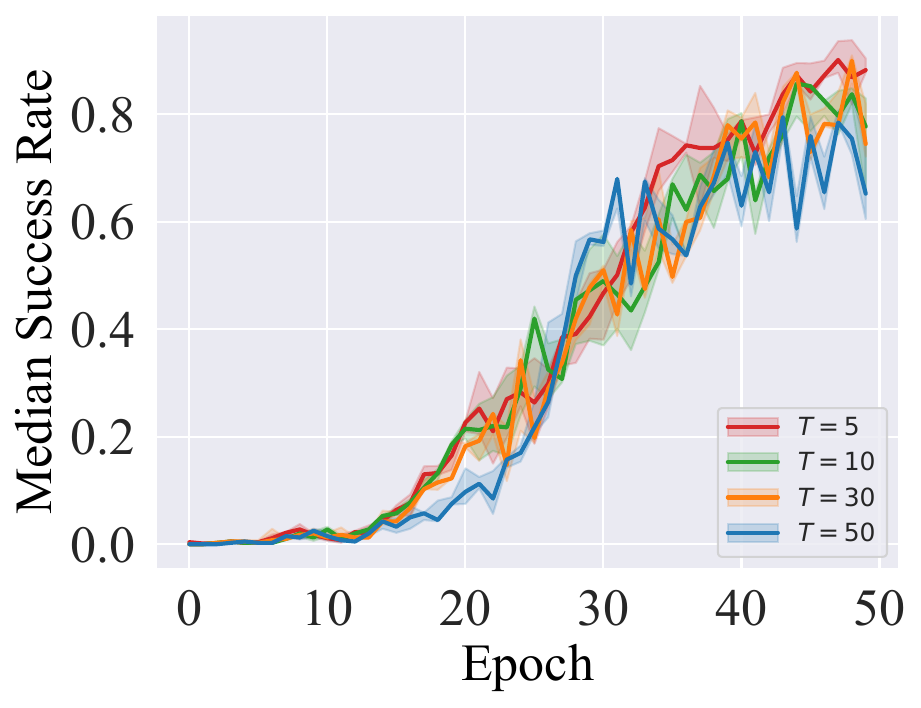}
        \caption{FetchSlide}
        \label{fig:ablation_fetch_slide}
	\end{subfigure}
	\caption{Through the ablation study, we analyze
the impact of different temperature parameters $T$ of sigmoid function on the success rate. The median success rate (line) and interquartile range (shaded) are shown with training 5 random seeds.}
       \label{fig:ablation_study}
\end{figure*}

\begin{algorithm}
    \caption{Contact Energy Based Prioritization~(CEBP)}\label{alg:cebp}
    \small{
    \begin{algorithmic}
        \State An off-policy algorithm $\mathbb{A}$ \Comment{In our case $\mathbb{A}$ is DDPG}
        \State $\mathcal{R} \leftarrow \emptyset$
          \For{$episode=1 , M$}
          \State Sample a desired goal $g$ and an initial state $s_0$
          \For{$t = 0 , \mathcal{T}-1$} \Comment{Rollout one episode}
          \State Collect a trajectory by running the policy  from $\mathbb{A}$ and save to the replay buffer $\mathcal{R}$;
             \State $a_t = \pi(s_t,g)$; 
             \State $r_t=r(s_t,a_t,g)$;
             \State $\mathcal{R}_i \leftarrow \left( s_t, a_t, r_t, s_{t+1}, g \right)$ \Comment{Store transition}
			\State $G:= \mathcal{S}(\text{current episode})$ \Comment{Sample a set of additional goals for replay}
				   \For{ $g' \in G$}\Comment{See HER\cite{AndrychowiczWRS17} }
 						\State $r'_t = r(s_t,a_t,g')$ \Comment{Recompute reward}
						\State $\mathcal{R}_i \leftarrow \left( s_t, a_t, r'_t, s_{t+1}, g' \right)$ \Comment{Store transition}
					\EndFor
          \EndFor
 		\State Compute $p_{\textrm{episode}}\left(e\right)$ using Eq.(\ref{eq:p_episode_based_on_contact})
 		\State Sample a minibatch $\mathcal{B}:$ $\{(s_t,a_t,s_{t+1},r_t,g)_i\}_{i}^{N} \sim \mathcal{R} $;
		\For{$j = 1$,  $k$}
			\State $(s_j,a_j,s_{j+1},r_j,g) \leftarrow \mathcal{B}(j)$
         	\State Compute IS weight:
                \State $w_j = \left( N \cdot p_{\textrm{episode}}(e)\right)^{- \beta} / \max(w)$ 
                \State Compute TD Error:
                \State $\delta_j = r_j + \gamma_{j} Q_{\textrm{target}} (s_j,g,\textrm{argmax}_a(Q' - Q))$ 
                \State Accumulate weight-change:
                \State $\Delta \leftarrow \Delta + w_j \delta_j \nabla_{\theta}Q$ 
		\EndFor
            \State Update weights
		\State $\theta \leftarrow \tau \cdot \theta$ + $\left( 1 - \tau \right) \cdot \theta'$
        \EndFor
    \end{algorithmic}}
    \label{alg:alg}
\end{algorithm}

\section{EXPERIMENT}
\label{sec:experiment}
This section describes the robot simulation benchmark used for evaluating the proposed approach. We test the proposed replay buffer using a 7-DOF robotic arm \cite{brockman2016openai} in the MuJoCo simulation \cite{todorov2012mujoco}. The robotic arm environment is designed as a standard benchmark for Multi-Goal RL \cite{plappert2018multigoal}. Three standard manipulation tasks from the benchmark, pick-and-place, push, and slide, are \textcolor{black}{ shown in Fig.\ref{fig:benchmark_env} and cloned from CPER where the original goal space of HER (green) is extended to a larger area (blue) in order to increase the difficulty of the tasks. Furthermore, the goals in the \textit{FetchPickAndPlace} are uniformly sampled to further increase the task difficulty (instead of using 50\% of targets on the table as in the HER benchmark task.)}

The state vector consists of the goal position as well as the position, orientation, linear and angular velocity of the end-effector and the object.
A goal represents the desired position for an object.
It is assumed that the task is accomplished if the object reaches the goal within the distance set by a threshold (i.e., defined as in equation~\eqref{eqn:reward_function}).
If the object is not in the range of the goal, the agent receives a negative reward $-1$; otherwise, the reward is $0$.
We compare the proposed approach CEBP, with 5 different random seeds, against other methods, namely CPER \cite{Vulin_2021}, PER~\cite{schaul2015prioritized}, MEP~\cite{zhao2019maximum}, EBP~\cite{zhao2018energy} and HER~ \cite{AndrychowiczWRS17} using three robotic manipulation tasks shown in Fig. \ref{fig:benchmark_env}.

\begin{itemize}[leftmargin=*]
\item Pick \& Place Task (\textit{FetchPickAndPlace-v1}):
The goal is to grasp the black object with a gripper and place it at the target location shown as a red dot in Fig. \ref{fig:fetch_pick_and_place}. The target positions are sampled from the green shaded region.
\item Push Task (\textit{FetchPush-v1}):
The position of both the target and the black object are sampled from the blue area on the table, and the robot's gripper is clamped as illustrated in Fig.~\ref{fig:fetch_push}. The goal is to push the object to the target position.
  \item Slide Task (\textit{FetchSlide-v1}):
The black object is sampled and placed on a long slippery table and the target is sampled from the blue area and it is outside of the robot's workspace (Fig.~\ref{fig:fetch_slide}). The goal is to hit the puck with a specific force so that it slides on the table and reaches the target location.
\end{itemize}




The outcomes of the simulation shown in Fig. \ref{fig:result_benchmark_env} provide insight into the efficacy of the contact energy-based prioritization~(CEBP) that has been proposed.
The agent, trained with the CEBP method, consistently outperforms existing benchmarks in both pick-and-place and push tasks. The agent converges to a near-optimal policy after $20$ epochs and $16$ epochs for the pick-and-place (Fig.\ref{fig:result_fetch_pick_and_place}) and push tasks (Fig.\ref{fig:result_fetch_push}), respectively. Especially, in the push task, our approach quickly finds out optimum actions at the early stages of training. In the slide task (Fig.\ref{fig:result_fetch_slide}), CPER initially starts to learn faster, but as the training process approaches the final phases, our approach achieves a similar final performance with a significantly lower standard deviation than CPER.
On the other hand, EBP considers only the object's energy and exhibits suboptimal performance compared to CEBP, as the tasks require object manipulation by the robot. The experimental results demonstrate that considering both object displacement and contact forces together for prioritizing the experiences results in better performance.
Furthermore, prioritizing experiences based on TD-error (PER) and a distribution with higher entropy (MEP) results in suboptimal outcomes. Notably, uniform prioritization of experiences (HER) exhibits the poorest performance among the considered methods.


When we analyzed the contact forces during the pick-and-place task, we observed that their magnitudes remained relatively constant when the object was being moved within the gripper towards the target goal. 
However, variations in the contact forces are observed during the picking up and the placing of the object. 
Consequently, the difference between using continuous contact forces (as in our method) and only two discrete values (as in CPER) for prioritizing experience appears to yield marginal differences, as shown in Fig.~\ref{fig:result_fetch_pick_and_place}.
However, in the push task, the robot pushes the object, which generates a wider range of contact forces than the other tasks.
Therefore, considering contact forces with a continuous range for prioritizing experience can make a significant difference in the learning rate compared to CPER, as shown in Fig. \ref{fig:result_fetch_push}.
In the slide task, contact between the robot and the object occurs for a very short time, resulting in limited contact feedback, and our method performs on par with the benchmarks.

The suboptimal performance of CPER \cite{Vulin_2021} can be attributed to the prioritization strategy.
During the prioritization in the CPER, the force values are accumulated through the timesteps, and when this cumulative sum surpasses a predetermined threshold, the corresponding forces are given a constant value of $\lambda$; otherwise, they are set to $1$.
As a result, a wide range of force values is compressed into a set with only two elements, $\{1, \lambda\}$.
Consequently, CPER does not consider varying levels of importance of the different force values for prioritization of the experiences.
In contrast to the CPER, we prioritize the experiences based on their wide range of contact-rich information using the sigmoid function.

In the ablation study, we investigate and compare the impact of temperature parameter $T$ of the sigmoid function $ \sigma(x) = \frac{k}{1 + e^{-xT}}$ on the performance where $k$ is the scale and $T$ is the temperature parameter.
We empirically found that parameters $k=100$ and $T=0.01$ for pick-and-place and push tasks and $k=100$ and $T=5$ for slide task lead to the best performance.
The aforementioned issue in CPER can be confirmed and inferred from the training outcomes in the ablation study, illustrated in Fig. \ref{fig:ablation_study}. 
If we increase the temperature parameter $T$ of the sigmoid function, the curve becomes much flatter, less steeper and less sensitive to contact changes and starts to behave like a binary switch and becomes gradually similar to the approach as in CPER. Accordingly, the success rate is decreasing based on the sensitivity in the sigmoid curve.

\section{Sim2Real}
\begin{figure}
    \centering
    \begin{subfigure}[b]{0.22\textwidth}
        \centering
        \includegraphics[width=\textwidth]{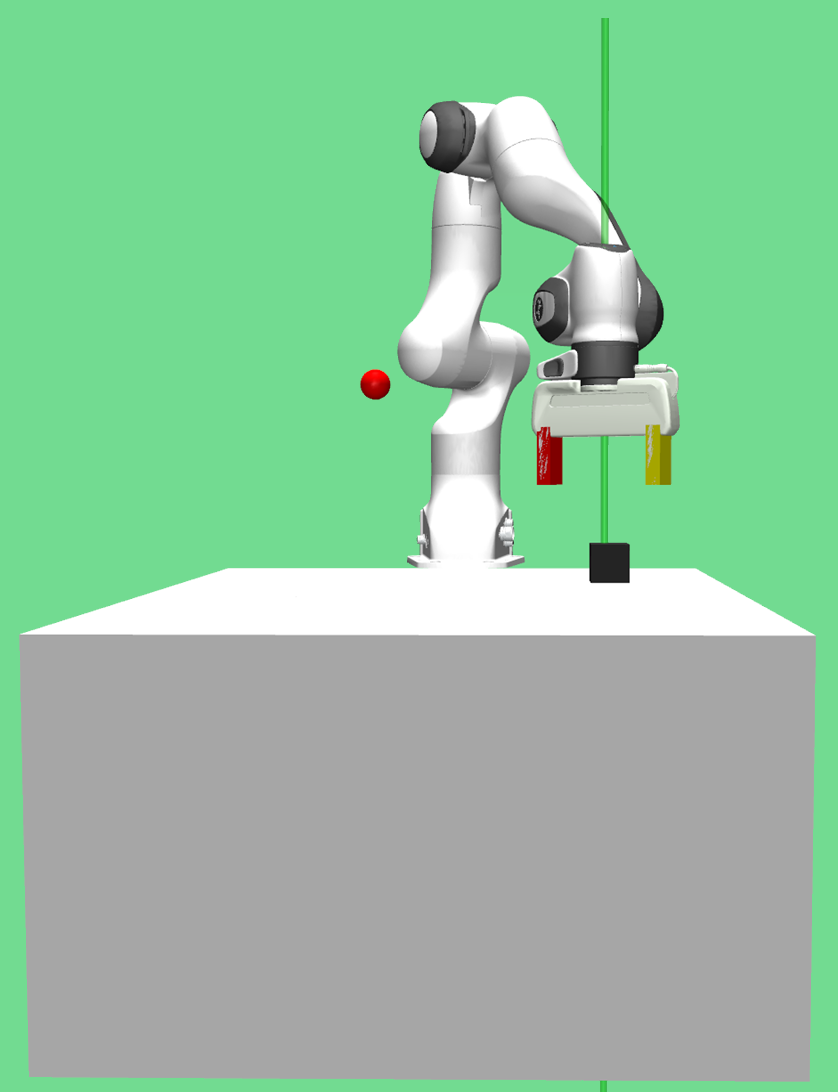}
        \caption{Franka Robot simulation}
        \label{fig:frankarobot_mujoco_simulation}
    \end{subfigure}
    \begin{subfigure}[b]{0.22\textwidth}
        \centering
        \includegraphics[width=\textwidth]{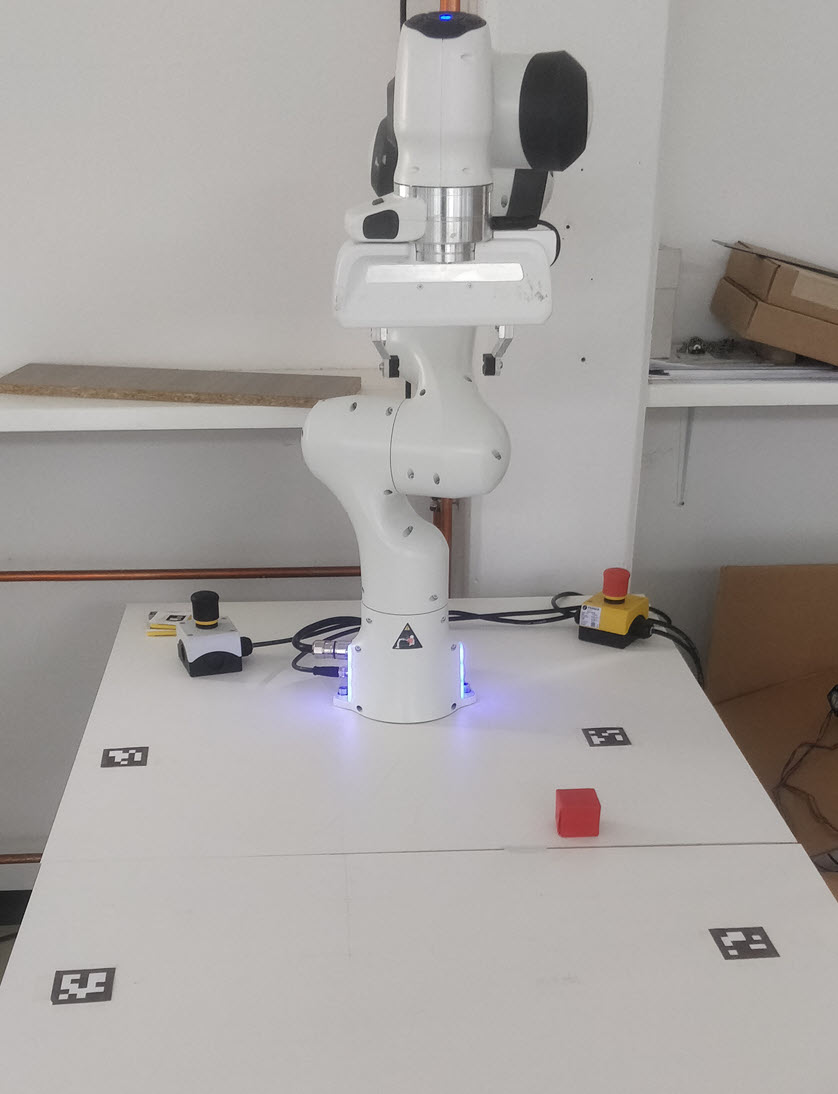}
        \caption{Franka Robot real setup}
        \label{fig:real_franka_robot}
    \end{subfigure}
    \caption{Franka robot setups}
       \label{fig:franka_robot_setups}
\end{figure}

\begin{figure}
    \centering
    \includegraphics[scale=0.4]{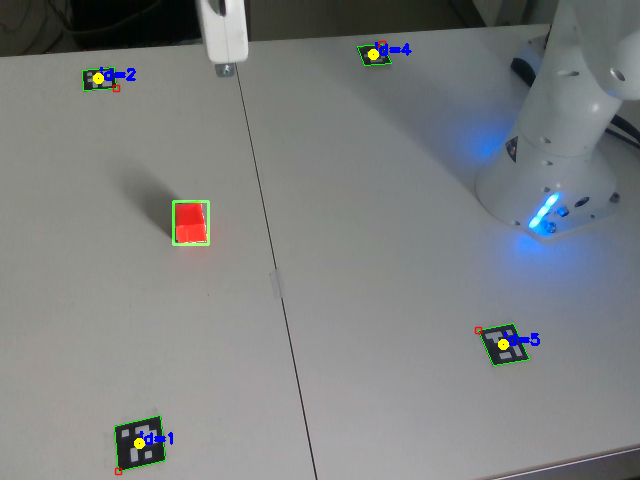}
    \caption{The camera view, showing the identified ArUco markers and detected object.}
    \label{fig:sim2real_from_camera_view}
\end{figure}

We trained the Franka robot in the MuJoCo simulation environment using a contact energy-based prioritization, shown in Fig.~\ref{fig:franka_robot_setups}.
The experiment was set up with a camera and ArUco markers.
ArUco markers and the red cube object are detected using the Aruco module and a red filter with the Open-CV library, respectively, and the centers of their pixel positions are obtained as shown in Fig.~\ref{fig:sim2real_from_camera_view}. The policy that has been trained is directly transferred from simulation to the real-world setting and outputs the linear motion in Cartesian space relative to its end-effector position as well as the state of the gripper gap. The parameters of linear motion are given to the Franka robot. Hereby, we robustly transferred the policy trained from the simulation environment into the real world.\

\section{Conclusions}
Limitations we faced during the Sim2Real transfer are the occlusion for the push task and a friction parameter discrepancy between simulation and real world settings in the slide task. Therefore, we were not able to execute the policy robustly in the real world setting in these tasks. However, we believe that the occlusion and friction problems can be overcome using multiple cameras from different perspectives and estimating the real world friction parameter, and using it in the simulation environment, respectively.

Our study introduces a new prioritization scheme named \textit{Contact Energy Based Prioritization}~(CEBP), which is designed for multi-goal RL with sparse rewards. We evaluated the efficacy of this approach in the MuJoCo simulation environment, testing it across three different robotic manipulation tasks. Experimental results indicate that this contact energy-based prioritization not only enhances performance but also improves exploration across three fundamental manipulation tasks.
Furthermore, the ablation study investigates the effect of the temperature parameter of the sigmoid function on the success rate. The suboptimal performance of the benchmark CPER is confirmed by the training results in the ablation study.
Additionally, we selected the pick-and-place task for Sim2Real transfer using a Franka robot. Based on our analysis, we can infer that the utilization of tactile feedback in conjunction with object displacement could serve as a catalyst for enhancing reinforcement learning performance, especially for contact-rich robot manipulation tasks.

For future work, we aim to investigate applying the concept of sparse reward and hindsight experience into meta-reinforcement learning (meta-RL) agents\cite{bing2023meta,bing2023meta2,wang2023meta,yao2023learning,bing2022meta3,zhou2023language}.

\section{Acknowledgements}
This research has been supported by the European Union’s Horizon 2020 Framework Programme for
Research and Innovation under the Specific Grant Agreement No.945539 (Human Brain Project SGA3) and German Federal Ministry of Education and Research (BMBF) (Project: 01IS22078). This work was also funded by Hessian.ai through the project `The Third Wave of Artificial Intelligence – 3AI` by the Ministry for Science and Arts of the state of Hessen.


\balance
\bibliographystyle{IEEEtran}
\bibliography{references}

\end{document}